\documentclass[review]{elsarticle}
\graphicspath{ {./figures/} }
\usepackage{hyperref}
\usepackage{float}
\usepackage{verbatim} 
\usepackage{apalike}
\usepackage{xcolor}
\usepackage{multirow,graphicx}
\usepackage{booktabs}
\usepackage{amssymb}
\restylefloat{figure}
\restylefloat{table}

\journal{Expert Systems with Applications}

\biboptions{authoryear}

\begin{document}

\newpage

\begin{center}
\LARGE Predicting student performance using sequence classification with time-based windows
 \newline

\large
\textbf{Galina Deeva, Johannes De Smedt, Cecilia Saint-Pierre, Richard Weber, Jochen De Weerdt}

Journal article (Accepted manuscript*)
\newline

\textbf{Please cite this article as:}

Deeva, G., De Smedt, J., Saint-Pierre, C., Weber, R., \& De Weerdt, J. (2022). Predicting student performance using sequence classification with time-based windows. Expert Systems with Applications, 118182. https://doi.org/10.48550/arXiv.2208.07749 \newline

DOI: \url{10.48550/arXiv.2208.07749} \newline

Available online 28 July 2022 

\end{center}

\newpage

\begin{frontmatter}










\title{Predicting student performance using sequence classification with time-based windows}

\author[label1]{Galina Deeva \corref{cor1}}
\ead{galina.deeva@kuleuven.be}

\author[label1]{Johannes De Smedt}
\ead{johannes.desmedt@kuleuven.be}

\author[label2]{Cecilia Saint-Pierre}
\ead{csaintpierre@uchile.cl}

\author[label3]{Richard Weber}
\ead{richard.weber@uchile.cl}

\author[label1]{Jochen De Weerdt}
\ead{jochen.deweerdt@kuleuven.be}

\cortext[cor1]{Corresponding author.}
\address[label1]{Research Centre for Information Systems Engineering, KU Leuven, Belgium}
\address[label2]{Online Learning Office, VTI, Universidad de Chile}
\address[label3]{Department of Industrial Engineering, FCFM,  Instituto Sistemas Complejos de Ingeniería, Universidad de Chile}

\begin{abstract}
A growing number of universities worldwide use various forms of online and blended learning as part of their academic curricula. Furthermore, the recent changes caused by the COVID-19 pandemic have led to a drastic increase in importance and ubiquity of online education. Among the major advantages of e-learning is not only improving students’ learning experience and widening their educational prospects, but also an opportunity to gain insights into students’ learning processes with learning analytics. This study contributes to the topic of improving and understanding e-learning processes in the following ways. First, we demonstrate that accurate predictive models can be built based on sequential patterns derived from students’ behavioral data, which are able to identify underperforming students early in the course. Second, we investigate the specificity-generalisability trade-off in building such predictive models by investigating whether predictive models should be built for every course individually based on course-specific sequential patterns, or across several courses based on more general behavioral patterns. Finally, we present a methodology for capturing temporal aspects in behavioral data and analyze its influence on the predictive performance of the models. The results of our improved sequence classification technique are capable to predict student performance with high levels of accuracy, reaching 90\% for course-specific models.
\end{abstract}


\end{frontmatter}

\section{Introduction}
\label{introduction}

With the recent growth of available educational technologies, online and blended learning have become crucial components of educational processes worldwide. The COVID-19 pandemic has strengthened this tendency significantly. More and more universities implement online assignments and even complete course modules as part of their academic curricula. Besides improving students’ learning experience and widening their educational prospects, such online courses also offer an opportunity to gain insights into students’ learning processes applying learning analytics \citep{viberg2018current}. 

Different combinations of online and offline learning become more common within educational institutions, such as blended learning and flipped classroom approaches, in which different parts of the learning process are happening both online and offline, and hybrid learning approaches, where some students are present in-class, while others are joining the class online. Analyzing data from such courses is a challenging task, due to dealing with rather incomplete datasets in comparison with purely online education. The incompleteness of the data captured by blended learning platforms is attributed to the fact that a large portion of learning processes happens outside of the learning platform and thus is invisible to it. Nevertheless, a recent study of a blended learning course has indicated that it is possible to predict student performance in such a challenging setting using just behavioral data \citep{van2020predicting}. However, the accuracy of these predictions remains relatively low and can still be further improved. Since it is not always feasible to address the problem of such a challenging setup of incomplete datasets with a few data points on the data level (for example, by obtaining more data), the solution can come on the algorithm level - by improving and tailoring existing algorithms to extract more information from the available data.

The main goal of this paper is to demonstrate that sequential pattern mining can be used to build models that can  detect which students are likely to perform poorly early-on in the course, both in online and blended learning setups. In particular, this paper puts forward an extension to a standard sequence classification approach by allowing to tailor the derived sequential patterns to be discovered from absolute time windows. More specifically, we present a flexible method to pinpoint the timing of sequential patterns within a course's time span, which can be flexibly selected by teachers and thus can facilitate interpretability. Given the importance of supporting teachers to organize their guidance more effectively and proactively, predicting student performance early-on is valuable, and with our method, we can enhance teacher control over the analysis and modeling, which is an interesting feature when specific information about the course is known.

Another key contribution of this work 
is the fact that, without the explicit use of sensitive socio-demographical variables including gender, ethnicity, age, social status, etc., we are capable to build accurate models for failure prediction. By only working with behavioural data, we can maximally avoid strongly undesired biases related to these socio-demographical variables and/or self-reported data.  


The experimental evaluation is performed using data from nine SPOCs hosted at the Learning Management System (LMS) platform at Universidad de Chile (UChile), called EOL. Using data from SPOCs allows us to enlarge the understanding of how incomplete data of the blended learning setting can be used for obtaining accurate and useful predictions of student performance. Moreover, another reason for focusing on private online courses as opposed to MOOCs is the additional costs involved in the case of on-campus education. Given such costs, it is generally more valuable to predict the performance of students enrolled in the university and bounded to graduate in comparison with the students who freely enroll in a MOOC without always having a clear intention to complete it. 

The rest of the paper is organized as follows. First, Section \ref{sec:related} discusses the current state of predictive learning analytics and its two important aspects: types of data and algorithms used. Then, in Section \ref{sec:cac}, the proposed approach is compared to other sequence mining methods available in the literature, highlighting their differences and similarities, and outlining the contribution of this work. Next, Section \ref{sec:background} describes how sequence classification can be used in the learning analytics context. Then, in Section \ref{sec:setup}, the data and the results of the predictions are presented. Finally, Section \ref{sec:discussion} contains information regarding possible implications of our findings, and Section \ref{sec:conclusion} summarizes the study.

\section{Related Work}
\label{sec:related}

Learning analytics (LA) is an important area of research mainly addressing the problem of revealing valuable and applicable knowledge from educational data \citep{daud2017predicting}. Its main purpose is to understand how students learn, as well as to potentially improve their learning outcomes \citep{viberg2018current}. A recent systematic literature review of LA in higher education identified the prediction of student performance as a common task \citep{viberg2018current}. This task is typically approached with classification and regression methods \citep{viberg2018current}: to predict a category of the learning outcome such as ‘passed’ and ‘failed’, or to predict a final score, respectively.

The studies that are focused on predicting student performance can be analyzed from different angles, but generally, two main dimensions can be utilized for their categorization. The first dimension deals with the kind of data and, consequently, features being used. Some authors aim to predict student performance based on students’ demographic data, their personality, or other survey-based data; see, e.g., \cite{daud2017predicting}.  Among other data which contributed to predicting student success are cumulative grade point average (GPA), assignments and quiz marks \citep{papamitsiou2014temporal}, past performances in previous assessments and entry tests, and demographics \citep{leitner2017learning}. \cite{shahiri2015review} broadly discussed the factors that contribute to the prediction of students’ performance.

Recently, data that represents the interactions of students with a learning system have been used in predictive analytics. \cite{agudo2014can} analyzed six online and two face-to-face courses to investigate whether students’ interaction data can be exploited to predict academic performance. Not unexpectedly, they show that this is possible only in online and not in face-to-face courses. Other studies that use behavioral data from online learning platforms can be found in, e.g., \cite{waheed2020predicting}, \cite{romero2013web} and \cite{popescu2018predicting}.

The second dimension for a categorization of predictive learning analytics research is the data analysis technique being used. As discussed above, various supervised machine learning approaches, such as classification and regression, are widely used, but other techniques are also observed in the literature. For instance, \cite{daud2017predicting} built five predictive models based on different algorithms, namely SVMs, CART, C4.5, Bayesian Networks, and Naive Bayes. Additionally, the authors of this study provide an overview of the most typical data mining methods used in predictive learning analytics. Besides the aforementioned techniques, the methods that are listed are neural network models, induction rules, association rules, and genetic algorithms, among others.

The actual process of feature engineering based on behavioral data is rarely discussed in detail, even though it is an important and challenging step. In the literature, it is common to observe relatively handcrafted and generally expert-driven features, i.e., proposed by the teacher and manually calculated rather than automatically discovered. As a result, such features could be rather simplistic in nature. For example, some of the features extracted from behavioral data by \cite{minaei2003predicting} are the total number of correct answers, the total number of tries for doing homework, the total time spent on the problem, etc. Similarly, \cite{waheed2020predicting} provide another example of constructing handcrafted features, such as the total number of clicks on a particular module, the number of attempts, and the number of assignments submitted.

The common approach to featurize behavioral data with activity (click) counts is, however, not always capable to capture the information regarding the order between the clicks and other temporal aspects. Sequence mining presents itself as a capable technique for featurisation of behavioral data because of its ability to capture the ordinal relationship between items in a sequence. Examples of sequence mining used in education can be found in studies by \cite{kinnebrew2013contextualized}, \cite{kotsiantis2004predicting}, and \cite{deeva2017dropout}. This study continues the work on this topic to discover better and smarter predictive and featurisation algorithms for predicting student performance.

\section{Comparative Analysis and Contributions}
\label{sec:cac}
In this section, we provide an overview of the most common sequence mining techniques as well as their key differences so as to position our approach. Table \ref{tab:comparison} depicts a comparative assessment in terms of four key dimensions which are both common in the sequence mining literature, as well as indicate the points of differentiation of the technique used in this work: (1) the type of algorithm, (2) the algorithm being support-based, (3) the pattern type used, and (4) the use of windows. We denote our approach as iBCM*, given that it is an adaptation of the original iBCM algorithm \citep{de2019mining}. 

First, sequence classification can be achieved in two ways: indirect or direct. For the former, a regular unsupervised sequential pattern mining algorithm is applied with the derived patterns subsequently being provided to a classification algorithm. The latter techniques make use of the label information to discover the most discriminating sequential patterns first, before similarly applying a classification algorithm of choice. iBCM* is an indirect algorithm, however, as pointed out by \cite{de2019mining}, it is outperforming direct algorithms such as BIDE\textsubscript{D(C)} \citep{fradkin2015mining}, MiSeRe \citep{DBLP:conf/icdm/EghoGBVC15}, and ISM \citep{masseglia2003incremental}, in terms of classification performance on a wide range of sequence classification datasets. 

 \begin{table}[htb]
 \centering
 \begin{tabular}{ll|llllllllll|l|}
 \toprule
 & & \parbox[t]{2mm}{\rotatebox[origin=c]{90}{PrefixSpan}}              & \parbox[t]{2mm}{\rotatebox[origin=c]{90}{SPADE}}  & 
     \parbox[t]{2mm}{\rotatebox[origin=c]{90}{cSPADE}}                  & \parbox[t]{2mm}{\rotatebox[origin=c]{90}{BIDE}}   & 
     \parbox[t]{2mm}{\rotatebox[origin=c]{90}{BIDE\textsubscript{D(C)}}}  & \parbox[t]{2mm}{\rotatebox[origin=c]{90}{MiSeRe}} & 
     \parbox[t]{2mm}{\rotatebox[origin=c]{90}{SCIS}}                    & \parbox[t]{2mm}{\rotatebox[origin=c]{90}{ISM}} & 
     \parbox[t]{2mm}{\rotatebox[origin=c]{90}{GoKrimp}}                 & \parbox[t]{2mm}{\rotatebox[origin=c]{90}{iBCM}} & 
     \parbox[t]{2mm}{\rotatebox[origin=c]{90}{iBCM*}}   \\
 \midrule
 \multirow{2}{*}{Type} & Direct   &            &            &            &            & \checkmark & \checkmark &            & \checkmark
                                  &            &            &            \\
                       & Indirect & \checkmark & \checkmark & \checkmark & \checkmark &            &            & \checkmark &
                                  & \checkmark & \checkmark & \checkmark \\
 \midrule
 \multicolumn{2}{c|}{Support-based}   & \checkmark & \checkmark & \checkmark & \checkmark & \checkmark &            & \checkmark &  \checkmark 
                                  & \checkmark & \checkmark & \checkmark \\
 \midrule
 \multirow{2}{*}{\shortstack{Pattern\\ type}} & Partial orders   & \checkmark & \checkmark &            & \checkmark & \checkmark & \checkmark &            
                                                  & \checkmark & \checkmark &            &            \\
                               & Constraints      &            &            & \checkmark &            &            &            &            
                                                  &            &            & \checkmark & \checkmark \\
 \midrule
 \multirow{2}{*}{Windows} & Relative length &            &             &            &            &            &            &        
                                           &            &            & \checkmark &            \\
                          & Absolute time   &            &            &            &            &            &            &            
                                           &            &            &            &  \checkmark \\
 \bottomrule
 \end{tabular}
 \caption{\label{tab:comparison}Table illustrating the key differences between iBCM* (our approach) and alternative sequence mining techniques.}
 \end{table}

Second, Table \ref{tab:comparison} shows a strong consistency in the support-based nature of a large majority of techniques. MiSeRe is the only exception, as it is essentially a parameter-free algorithm relying on a Bayesian reconstruction approach. The support-based nature of iBCM* is important because in the application to behavioral learning data, we require a flexible and tunable technique, given that the sequence data can be of very different granularity levels and that differentiating patterns in terms of student success can be complex. This is demonstrated in our experimental evaluation in Section \ref{sec:setup}.

More relevant for pinpointing the contribution of this work, is the comparison in terms of pattern types and windows. A majority of sequence mining techniques rely on partial orders (e.g. GoKrimp \citep{lam2014mining}, SCIS \citep{zhou2015pattern}, BIDE \citep{wang2004bide}, and SPADE \citep{zaki2001spade}). However, constraint-based approaches such as cSPADE \citep{zaki2000sequence} and iBCM allow for more flexible and richer patterns to be derived from the sequences. While cSPADE allows for typical constraints such as gap, item, and time constraints, iBCM relies on a comprehensive set of declarative behavioral templates, as further detailed in Section \ref{subsec:iBCM}, which intrinsically cover gap and item constraints. In \citep{de2019mining}, it was shown that this richness in derived sequential patterns has a positive impact on classification performance as well as the conciseness of the retrieved pattern base. As for student success prediction, the increased diversity in possible sequential patterns is desirable, given that a typical dataset usually contains relatively few observations, especially compared to sequence databases in other fields such as biology, phyiscs, web mining, or text mining. 
Nevertheless, iBCM has also been used and benchmarked on smaller sequence database \cite{de2019mining} as well as in the context of session stitching and web mining of large web logs \cite{de2021session}, indicating that these constraints serve predictive purposes in a multitude of contexts.

Finally, the last comparison dimension relates to window configuration and time. While many sequence mining algorithms could be configured to do something similar, iBCM is the only technique that can natively work with windows. This entails that iBCM can be restricted to discover sequential patterns from subsequences defined based on their relative length. For instance, an original complete sequence with 20 items, can be subdivided into 4 subsequences of 5 items, with patterns only derived within each window. The core contribution of this paper is that we extend this window-based approach to absolute time windows. This is particularly relevant for characterizing sequences pertaining to learning processes by students, given that time is a crucial factor in the characterization of how students engage and learn. 

In conclusion, the core contribution of this work is a learning analytics-oriented extension of iBCM to allow for the construction of student success prediction models based on behavioral data. By combining the comprehensive set of constraint types of iBCM with custom absolute time-based windows, we can better capture the temporal aspects affecting student success.

\section{Predicting student performance using sequence classification: methodology}
\label{sec:background}
In this section, the core methodology for building predictive models based on sequence classification is presented. We use the iBCM (interesting Behavioral Constraint Miner) sequence classification algorithm introduced by \cite{de2019mining} for deriving sequential patterns. iBCM is a state-of-the-art sequence classification technique supporting a wide range of pattern types, which makes it a particularly suitable technique for adapting it to a specific prediction task. iBCM can be used for featurising learners' activity sequences and requires to be paired with a classification algorithm to construct a predictive model.

In what follows, behavioral data in the context of success prediction is discussed. Then, the way sequential patterns are extracted is outlined and followed by an explanation of how to build a predictive model. Finally, an adaptation of the baseline sequence classification technique is introduced that can capture more specific learning-driven temporal aspects in sequences through time-based windows. 

\subsection{Analyzing behavioral data from an e-learning platform}
In the context of online learning, one of the most easily available and unambiguous sources of student data is the student activity recorded by an e-learning platform. Other types of data exist as well and could potentially provide valuable information for predicting student performance, e.g., demographic data such as age and previous education, students’ affective states and their transitions over time, students’ personality characteristics such as level of neuroticism and achievement striving. However, these types of data are not always available for all students and/or might suffer from reporting bias. Moreover, demographic data can easily lead to biased models in terms of gender, ethnical characteristics, etc. By only using behavioral data, we aim at building models that are as unbiased as possible with respect to personal characteristics.

\begin{figure}[!ht]
	\centering
		\includegraphics[width=1\textwidth]{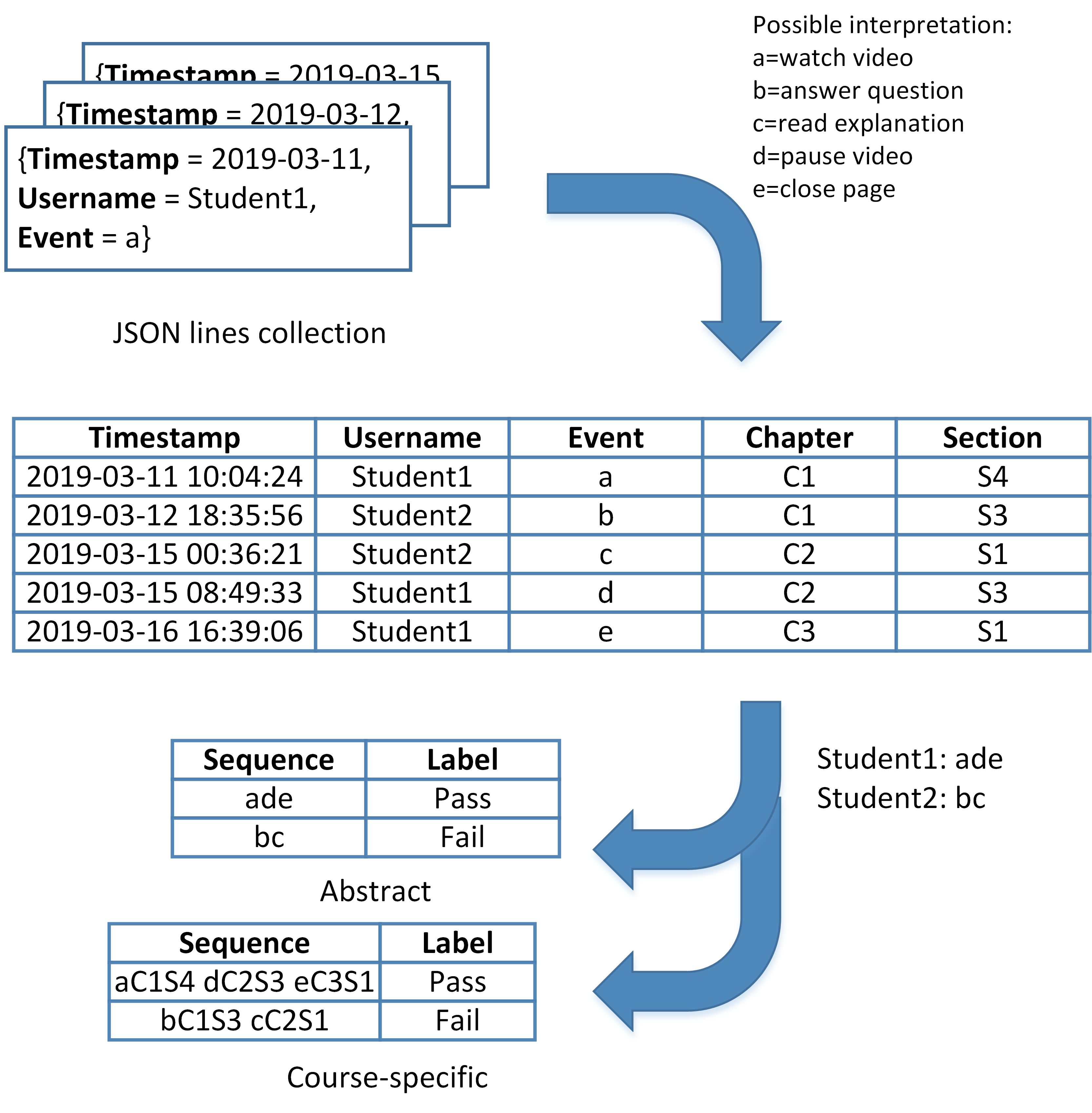}
		\caption{An example of behavioral data and its transformations needed to perform a sequential data analysis}
		\label{dataformat}
\end{figure}

Tracking data, also often referred to as behavioral data, represents student behavior captured by an e-learning platform. In this paper, we analyze the data from the EOL platform, the edX-based LMS used in UChile to run courses for undergraduate and graduate schools as well as professional education and MOOCs. This LMS provides tracking data in a streaming newline-delimited JSON (NDJSON) format. More specifically, the data in NDJSON format is a collection of lines, each of which is a valid nested JSON structure. These data can be transformed into a sequence as illustrated in Figure \ref{dataformat}. Firstly, it can be flattened and unnested to obtain tabular data, containing timestamp, username, event, and its context, i.e. chapter and section of the course. Generally, there are several possible ways to obtain a sequence from such data. As shown by \cite{deeva2017dropout}, three levels of event granularity can be considered: the most general level with event information only, an intermediate level in which the event name is combined with the chapter information and, finally, the most fine-granular level in which event names are combined with chapter and section. 

Based on a chosen activity granularity level, one can obtain sequences of events, in sequence mining also called as sequences of items. In the remainder of this paper, “event” and “item” are used to refer to the recording of a student action within the platform, characterized by a particular timestamp.

\subsection{Mining sequential patterns from behavioral data using iBCM}
\label{subsec:iBCM}
The iBCM algorithm uses label information to derive the most discriminative patterns for a certain class of sequences. These patterns are then used as binary features indicating a presence or absence of a pattern, which allows to transform a set of sequences into a feature matrix, as shown in Figure 2. iBCM has been previously shown to achieve high levels of performance in terms of model accuracy in a relatively short computational time \citep{de2019mining}.

One of the key advantages of iBCM is the richness of the sequential patterns that it supports, as it combines higher expressiveness compared to partial orders with a hierarchy of sequential behavior ranging from a simple order to a next-in-sequence and alternating behavior. Some of the other sequence mining algorithms, e.g. BIDE and PrefixSpan, use partial orders, while iBCM is capable to exploit a whole hierarchy of sequential patterns. The sequential patterns that can be discovered by iBCM are based on a set of Declare language templates \citep{di2013two}. These patterns are widely used for identifying not only sequential, but also overall behavioral characteristics of programs and processes. The Declare template base consists of a number of patterns for modeling flexible business processes, and it is extensive. The set of templates that is used by the current version of iBCM is listed in Table \ref{tab:templates}.

\begin{table*}[!ht]
\centering
	\caption{An overview of behavioral templates mined by iBCM and their corresponding regular expressions}
		\begin{tabular}{lll}
		
			\hline
			\textbf{Type} & \textbf{Template}  & \textbf{Regular Expression}  \tabularnewline
			\hline
			Unary &
			Existence(A,n)  & .{*}(A.{*})\{n\} \tabularnewline
			& Absence(A,n) & {[}\^{ }A{]}{*}(A?{[}\^{ }A{]}{*})\{n-1\}  \tabularnewline
			& Exactly(A,n) &{[}\^{ }A{]}{*}(A{[}\^{ }A{]}{*})\{n\} \tabularnewline
			& Init(A) & (A.{*})? \tabularnewline
			& Last(A)  & .{*}A \tabularnewline\hline 
		Unordered &
			Responded existence(A,B) & {[}\^{ }A{]}{*}((A.{*}B.{*}) $|$(B.{*}A.{*}))?\tabularnewline
			& Co-existence(A,B) & {[}\^{ }AB{]}{*}((A.{*}B.{*}) $|$(B.{*}A.{*}))?\tabularnewline\hline
		Simple ordered&
			Response(A,B)  & {[}\^{ }A{]}{*}(A.{*}B){*}{[}\^{ }A{]}{*}   \tabularnewline
			& Precedence(A,B) & {[}\^{ }B{]}{*}(A.{*}B){*}{[}\^{ }B{]}{*} \tabularnewline
			& Succession(A,B) & {[}\^{ }AB{]}{*}(A.{*}B){*}{[}\^{ }AB{]}{*} \tabularnewline\hline
			Alternating&
			Alternate response(A,B)  & {[}\^{ }A{]}{*}(A{[}\^{ }A{]}{*}B{[}\^{ }A{]}{*}){*} \tabularnewline
			ordered &Alternate precedence(A,B) &  {[}\^{ }B{]}{*}(A{[}\^{ }B{]}{*}B{[}\^{ }B{]}{*}){*}  \tabularnewline
			& Alternate succession(A,B) & {[}\^{ }AB{]}{*}(A{[}\^{ }AB{]}{*}B{[}\^{ }AB{]}{*}){*} \tabularnewline\hline
		Chain ordered&
			Chain response(A,B) & {[}\^{ }A{]}{*}(AB{[}\^{ }A{]}{*}){*} \tabularnewline
			& Chain precedence(A,B) &  {[}\^{ }B{]}{*}(AB{[}\^{ }B{]}{*}){*} \tabularnewline
			& Chain succession(A,B) & {[}\^{ }AB{]}{*}(AB{[}\^{ }AB{]}{*}){*} \tabularnewline\hline
			Negative&
			Not co-existence(A,B)  & {[}\^{ }AB{]}{*}((A{[}\^{ }B{]}{*}) $|$(B{[}\^{ }A{]}{*}))?  \tabularnewline
			& Not succession(A,B) & {[}\^{ }A{]}{*}(A{[}\^{ }B{]}{*}){*}  \tabularnewline
			& Not chain succession(A,B) & {[}\^{ }A{]}{*}(A+{[}\^{ }AB{]}{[}\^{ }A{]}{*}){*}A{*}  \tabularnewline\hline
			Choice&
			 Choice(A,B) &  .{*}{[}AB{]}.{*} \tabularnewline
			& Exclusive choice(A,B) &  ([\^{ }B]*A[\^{ }B]*) $|$.{*}{[}AB{]}.{*}({[}\^{ }A{]}{*}B{[}\^{ }A{]}{*}) \tabularnewline
			\hline
		\end{tabular}
	\label{tab:templates}
\end{table*}

\subsection{Building predictive models based on sequential patterns}
\begin{figure}[!ht]
	\centering
		\includegraphics[width=1\textwidth]{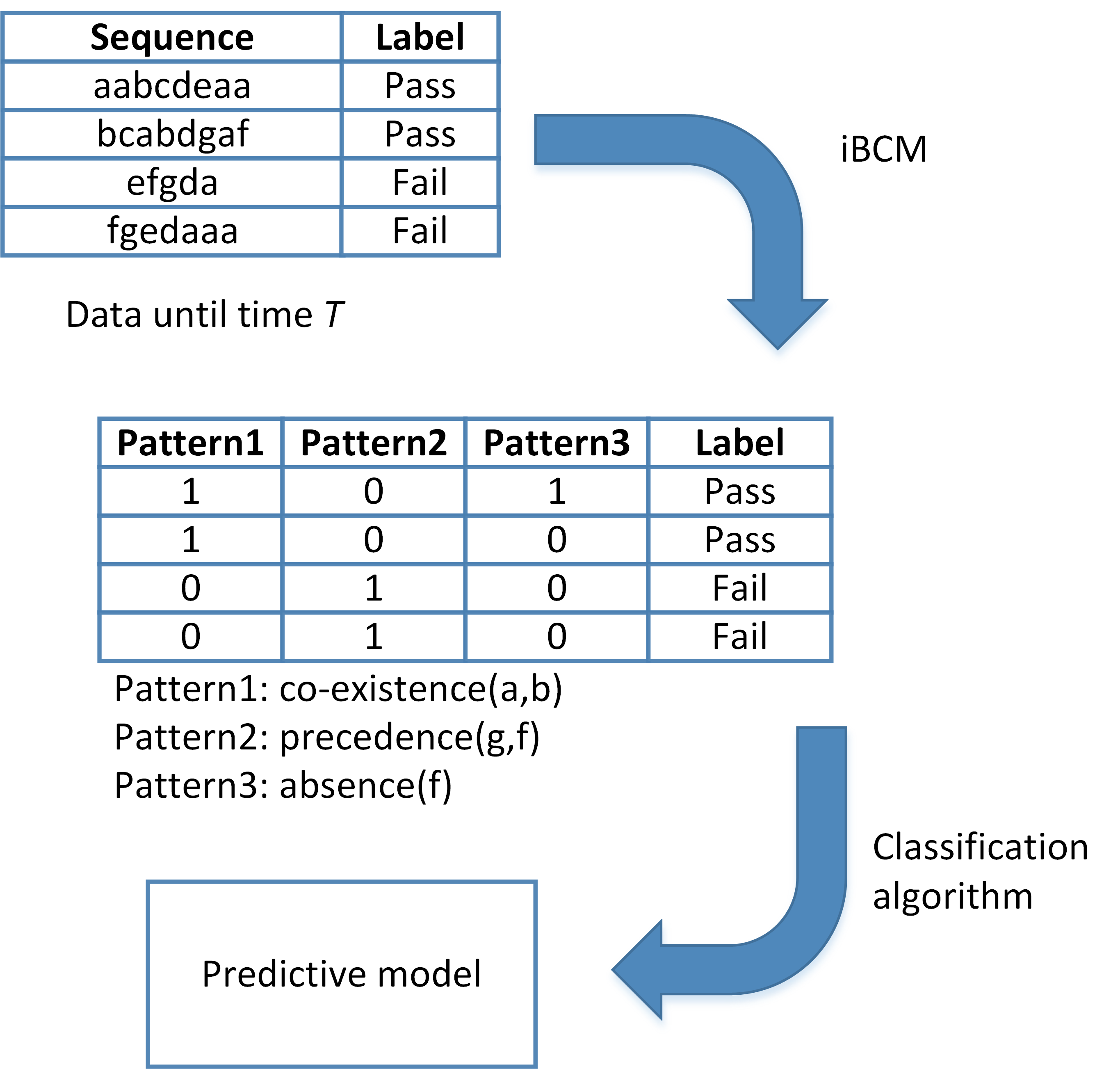}
		\caption{An illustration of the sequence classification process with iBCM}
		\label{iBCM}
\end{figure}

As illustrated in Figure \ref{iBCM}, after sequences are featurised, i.e., transformed into a feature matrix using the iBCM-derived patterns as features, a classification algorithm can be applied. There are several aspects that need to be considered when doing so. First of all, to prevent potential data leakage and to ensure actionability of the model, only the data (sequences) up to a certain time T can be used to predict student performance \citep{deeva2017dropout}. In terms of data leakage, using the data from the whole course might potentially contain cases of students that already dropped out from the course or simply engaged in much less active participation on the platform, which might bias the algorithm. In terms of actionability of the prediction, it is more relevant and practical to pinpoint potential students at-risk as early as possible during the course, as this could help the teachers to support these students before the course is finished. On the other hand, there is a trade-off between early prediction and the amount of data available for building a model. Therefore, it is important to choose a suitable moment \textit{T}, which would be an optimal point within the course for making the prediction. 

\subsection{Capturing temporal aspects}
\begin{figure}
	\centering
		\includegraphics[width=1\textwidth]{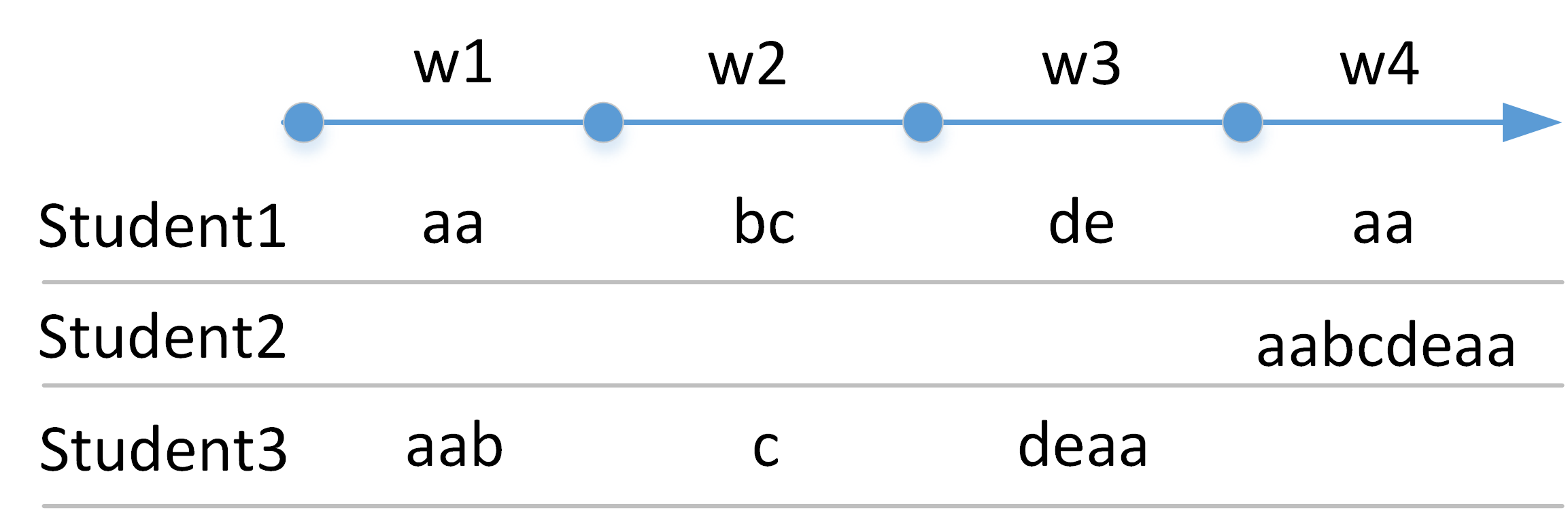}
		\caption{An example of the situation where the information regarding temporal aspect could be substantial for predicting student performance}
		\label{temporal_aspect}
\end{figure}

The advantage of sequence classification for analyzing temporal data in comparison with simpler featurisation methods, such as for example the bag-of-words approach where number of occurrences of each item is used as a feature, is that it can take into account not only the type of items in the sequences, but also their order. However, in case of educational data, there could be situations in which the order of items is not enough to differentiate between students with different performance levels. Figure \ref{temporal_aspect} shows an example that illustrates this idea using a four-week course. Each week could be considered as a time window. Three students have performed the same sequence of activities, namely $<aabcdeaa>$, and therefore will be treated way by a sequence classification algorithm and will be predicted to achieve the same level of performance. However, as is clear from the figure, the first student has been working consistently throughout the semester, whereas the second student has completed all the work at the last possible moment, i.e. within the last week/window. Given that previous research in learning analytics has shown that the way the students approach deadlines is correlated with their final results \citep{celis2015modelo}, we could potentially expect the first student to perform better than the second.

To capture this difference, we introduce a notion of time-based windows that can be implemented with iBCM. Currently iBCM already has a functionality that allows to pinpoint patterns that are found in a particular window. However, these windows are defined based on the number of items. More specifically, a complete sequence is divided into n windows of equal size, i.e., equifrequent windows, where $n$ is a user-defined parameter. This means that with $n=4$, a sequence $<aabcdeaa>$, which consists of eight items, will be divided into 4 subsequences of 2 items each. In case of educational data, this abstraction will facilitate yielding more fine-granular patterns, but still will not capture the difference between students that demonstrated different attitude in terms of time. Therefore, in this paper we propose to tailor iBCM to discover patterns based on equidistant windows by taking into account absolute times. As such, iBCM will be capable of picking up the temporal differences between students in the discovered patterns. While the equifrequent windows version of iBCM would discover the same set of patterns in the sequences of three students shown in Figure \ref{temporal_aspect}, the equidistant windows version would yield three different sets of behavioral patterns.
\section{Experimental evaluation}
\label{sec:setup}

\subsection{Data description}

This study analyses behavioral data from nine SPOCs, as summarized in Table 2. The models for three courses with the highest number of students are built, namely \textit{FCS-001}, \textit{ENFGF243}, and \textit{ENMAN303}.

In the scoring system used by these courses, 7 is the best score and 4 the worst score to pass. The relatively low percentage of students who failed a course (roughly 7\% for all courses combined, and 0\% for some of the courses) makes the prediction of student success even more challenging, given the data imbalance and small number of observations (i.e. students). To surpass this challenge, we aim at predicting the students who performed in the bottom 20 percent.

For each of these three courses, predictive models for general and course-specific patterns are constructed, with the goal of comparing their performance and thus investigating whether there is a significant difference between these two cases. 
Moreover, to assess the possibility of generalizing the patterns across multiple courses, another set of predictive models is built for all nine courses combined. This setting can only be applied in case of general patterns since course-specific patterns cannot be generalized for a combination of all the datasets. The results of the algorithm performance for all courses combined allow to draw conclusions not only regarding the usability of general patterns, but also whether such indicative patterns are common across different courses and can be generalisable.

Furthermore, the performance of iBCM with temporal windows is studied by comparing three setups. First, iBCM is applied in its default mode, i.e., patterns are captured in the whole sequence. Secondly, the time-based window approach is applied to obtain subsequences based on half of the analyzed timeline, so resulting in two equidistant windows. Finally, the approach is also applied so that subsequences are formed that are covering one fourth of the analyzed timeline, so obtaining four equidistant windows.

The data characteristics at course level are summarized in Table \ref{tab:eol_overview}. The first column represents the code of the course and its duration; the second column shows the total number of students ($N_T$).  
The distribution of the course grade is summarized in the third column, where the mean and the median of the courses’ grades are given. $Q_{20}$ is the grade of the student who scored in the $20^{th}$ percentile, which means that all the students who scored below this number are marked as the predicted target variable. $N_{20}$ is the number of students whose grade is below $Q_{20}$. Finally, the fourth column provides the total numbers of events per course (Total), as well as alphabet sizes for the general and course-specific event granularity ($N_{items}$).

\begin{table*}[!ht]
\caption{{Overview of the courses analyzed in this study, including their duration, score distributions, total number of events and alphabet sizes for the abstract and course-specific event granularity} }
\label{tab:eol_overview}
\def\arraystretch{1}
\ignorespaces 
\centering 
\scriptsize
\begin{tabular}{p{2.7cm}p{1.8cm}p{2.4cm}p{2.8cm}}
\hline Course \newline Duration & \# of students 
& Score  distribution
\newline 20th percentile & \# of events \newline Alphabet size \\
\hline 
FCS-001 \newline 22 weeks &
  $N_{T} = 247$ 
  &
  mean = 5.63 \mbox{}\protect\newline median = 5.90 \mbox{}\protect\newline Q\ensuremath{_{20}} = 5.00 \mbox{}\protect\newline N\ensuremath{_{20}} = 47 &
  Total = 49,151 \mbox{}\protect\newline N\ensuremath{_{items\ }}= 23 (201)\\
\hline 
ENFGF243 \mbox{}\protect\newline 16 weeks &
  N\ensuremath{_{T\ }}= 214 \mbox{}\protect
  &
  mean = 6.24 \mbox{}\protect\newline median = 6.33 \mbox{}\protect\newline Q\ensuremath{_{20}} = 6.04 \mbox{}\protect\newline N\ensuremath{_{20}} = 39 &
  Total = 110,758 \newline N\ensuremath{_{items\ }}= 24 (179)\\
\hline 
ENMAN303 \mbox{}\protect\newline 17 weeks &
  N\ensuremath{_{T\ }}= 147 \mbox{}\protect
  &
  mean = 5.78 \mbox{}\protect\newline median = 6.40 \mbox{}\protect\newline Q\ensuremath{_{20}} = 5.50 \mbox{}\protect\newline N\ensuremath{_{20}} = 28 &
  Total = 62,329 \mbox{}\protect\newline N\ensuremath{_{items\ }}= 28 (228)\\
\hline 
ENNEG388 \mbox{}\protect\newline 15 weeks &
  N\ensuremath{_{T\ }}= 94 \mbox{}\protect 
  &
  mean = 5.36 \mbox{}\protect\newline median = 5.70 \mbox{}\protect\newline Q\ensuremath{_{20}} = 4.80 \mbox{}\protect\newline N\ensuremath{_{20}} = 17 &
  Total = 44,809 \mbox{}\protect\newline N\ensuremath{_{items\ }}= 28\\
\hline 
CFG-TUT-V1 \mbox{}\protect\newline 20 weeks &
  N\ensuremath{_{T\ }}= 75 \mbox{}\protect
  &
  mean = 4.80 \mbox{}\protect\newline median = 5.90 \mbox{}\protect\newline Q\ensuremath{_{20}} = 4.60 \mbox{}\protect\newline N\ensuremath{_{20}} = 14 &
  Total = 8,818 \mbox{}\protect\newline N\ensuremath{_{items\ }}= 22\\
\hline 
CIE-115.418 \mbox{}\protect\newline 16 weeks &
  N\ensuremath{_{T\ }}= 60 \mbox{}\protect
  &
  mean = 5.43 \mbox{}\protect\newline median = 5.40 \mbox{}\protect\newline Q\ensuremath{_{20}} = 5.00 \mbox{}\protect\newline N\ensuremath{_{20}} = 11 &
  Total = 15,853 \mbox{}\protect\newline N\ensuremath{_{items\ }}= 19\\
\hline 
TUT-CFP-DMD \mbox{}\protect\newline 10 weeks &
  N\ensuremath{_{T\ }}= 24 \mbox{}\protect
  &
  Pass/Fail \mbox{}\protect\newline N\ensuremath{_{F(20)}} = 4 &
  Total = 1,566 \mbox{}\protect\newline N\ensuremath{_{items\ }}= 23\\
\hline 
ENFGF106 \mbox{}\protect\newline 17 weeks \mbox{}\protect\newline  &
  N\ensuremath{_{T}}  = 23 \mbox{}\protect
  &
  mean = 6.22 \mbox{}\protect\newline median = 6.20 \mbox{}\protect\newline Q\ensuremath{_{20}} = 5.90 \mbox{}\protect\newline N\ensuremath{_{20}} = 4 &
  Total = 13,696 \mbox{}\protect\newline N\ensuremath{_{items\ }}= 23\\
\hline 
CIE-FC720 \mbox{}\protect\newline 7 weeks \mbox{}\protect\newline  &
  N\ensuremath{_{T}}  = 14 \mbox{}\protect
  &
  mean = 5.49 \mbox{}\protect\newline median = 5.50 \mbox{}\protect\newline Q\ensuremath{_{20}} = 4.80 \mbox{}\protect\newline N\ensuremath{_{20}} = 2 &
  Total = 4,406 \mbox{}\protect\newline N\ensuremath{_{items\ }}= 19\\
\hline 
All courses &
  N\ensuremath{_{T}}  = 895 \mbox{}\protect
  \mbox{}\protect\newline  &
  $N_{20} = 163$ &
  Total = 311,386 \newline N\ensuremath{_{items\ }}= 30\\
\hline 
\end{tabular}\par 
\end{table*}

All experiments have been performed using four classifiers: Support Vector Machines (SVM), logistic regression, Random Forest, and Naïve Bayes. Random minority class oversampling is applied as a method for handling class imbalance for all the experiments. The evaluation of the models is performed using 10-fold cross-validation. Given that SVM has achieved the highest levels of performance on average, all results are reported using SVM. 

Finally, iBCM uses the hyperparameter called “support”, which is defined as a proportion of analyzed sequences which needs to contain the pattern for it to be considered frequent. For example, a support value of 0.6 implies that 60\% of sequences must contain a certain pattern to call it frequent. To eliminate potential influences of the support level on other investigated parameters, all results are calculated for three different support levels: 0.4, 0.6, and 0.8.

\subsection{Results}
The results of the experimental evaluation are summarized in Tables \ref{tab:results} and \ref{tab:all_courses}, using the following three performance metrics:

\begin{enumerate}

    \item Accuracy, i.e. the percentage of correct classifications;
    \item AUC (area under the ROC curve), which measures the trade-off between the true and false positive rates over different classification thresholds;
    \item Top decile lift (from here on: lift), which measures how much better a classification model is, compared to a random selection, considering the top 10\% predicted instances with the highest prediction probability.
    
\end{enumerate}

Overall, sequence classification with iBCM is able to achieve good performance, with the average accuracy, AUC, and lift for all experiments equal to 0.7, 0.65, and 2.46, respectively. Moreover, some of the constructed models achieve very high levels of performance, e.g., the model for course-specific patterns for the \textit{ENMAN303} course, where the metrics are $accuracy=0.91$, $AUC=0.9$, and $lift=5.03$ (see Table \ref{tab:results}). The reason the models built for this course were capable of achieving the best predictive performance among all the analyzed courses is that this is the only course in which the learning process occurs completely online, unlike the others which utilized a flipped classroom approach. Overall, these results provide additional evidence that sequence classification in general, and iBCM in particular, form a suitable and efficient tool for mining educational data. 

\begin{table*}[!htp]
\caption{{Overview of the results for the ENFGF243 and ENMAN303 courses. All experiments are performed using SVM and random oversampling of the minority class. The metrics are assessed by using 10-fold cross validation. The best result per metric for a combination of course, pattern type and support is shown in bold.} }
\label{tab:results}
\def\arraystretch{1}
\ignorespaces 
\centering 
\scriptsize
\begin{tabular}{llrrrrrr}
\hline
    Dataset & Patterns & \multicolumn{1}{l}{support} & \multicolumn{1}{l}{w} & \multicolumn{1}{l}{\#patterns} & \multicolumn{1}{l}{Acc} & \multicolumn{1}{l}{AUC} & \multicolumn{1}{l}{lift} \\
    \hline
    ENFGF243 & Abstract & 0.4   & 1     & 79    & 0.66  & 0.55  & \textbf{1.26} \\
    ENFGF243 & Abstract & 0.4   & 2     & 174   & \textbf{0.64} & \textbf{0.63} & \textbf{1.26} \\
    ENFGF243 & Abstract & 0.4   & 4     & 122   & \textbf{0.64} & 0.62  & \textbf{1.26} \\
    \hline
    ENFGF243 & Abstract & 0.6   & 1     & 3266  & \textbf{0.76} & \textbf{0.71} & \textbf{3.24} \\
    ENFGF243 & Abstract & 0.6   & 2     & 1370  & 0.75  & 0.66  & \textbf{3.24} \\
    ENFGF243 & Abstract & 0.6   & 4     & 840   & \textbf{0.76} & 0.61  & 3.02 \\
    \hline
    ENFGF243 & Abstract & 0.8   & 1     & 46    & \textbf{0.66} & \textbf{0.59} & \textbf{0.99} \\
    ENFGF243 & Abstract & 0.8   & 2     & 8     & 0.57  & 0.54  & \textbf{0.99} \\
    ENFGF243 & Abstract & 0.8   & 4     & 12    & 0.56  & 0.56  & \textbf{0.99} \\
    \hline
    ENFGF243 & Course-specific & 0.4   & 1     & 3900  & 0.75  & 0.67  & 2.74 \\
    ENFGF243 & Course-specific & 0.4   & 2     & 4055  & \textbf{0.79} & \textbf{0.70} & \textbf{3.51} \\
    ENFGF243 & Course-specific & 0.4   & 4     & 1473  & 0.76  & 0.56  & 3.24 \\
    \hline
    ENFGF243 & Course-specific & 0.6   & 1     & 3266  & \textbf{0.76} & \textbf{0.71} & \textbf{3.24} \\
    ENFGF243 & Course-specific & 0.6   & 2     & 1370  & 0.75  & 0.66  & \textbf{3.24} \\
    ENFGF243 & Course-specific & 0.6   & 4     & 840   & \textbf{0.76} & 0.61  & 3.02 \\
    \hline
    ENFGF243 & Course-specific & 0.8   & 1     & 1870  & 0.71  & 0.67  & 3.02 \\
    ENFGF243 & Course-specific & 0.8   & 2     & 475   & 0.70  & 0.63  & \textbf{3.24} \\
    ENFGF243 & Course-specific & 0.8   & 4     & 340   & \textbf{0.74} & \textbf{0.68} & \textbf{3.24} \\
    \hline
    ENMAN303 & Abstract & 0.4   & 1     & 131   & 0.77  & 0.71  & \textbf{3.15} \\
    ENMAN303 & Abstract & 0.4   & 2     & 298   & \textbf{0.79} & \textbf{0.82} & 2.87 \\
    ENMAN303 & Abstract & 0.4   & 4     & 186   & 0.78  & 0.73  & 2.96 \\
    \hline
    ENMAN303 & Abstract & 0.6   & 1     & 104   & 0.76  & 0.65  & 3.52 \\
    ENMAN303 & Abstract & 0.6   & 2     & 196   & \textbf{0.80} & \textbf{0.78} & 2.87 \\
    ENMAN303 & Abstract & 0.6   & 4     & 148   & 0.77  & 0.74  & \textbf{3.74} \\
    \hline
    ENMAN303 & Abstract & 0.8   & 1     & 258   & 0.72  & 0.64  & \textbf{3.15} \\
    ENMAN303 & Abstract & 0.8   & 2     & 211   & \textbf{0.75} & \textbf{0.80} & 2.87 \\
    ENMAN303 & Abstract & 0.8   & 4     & 193   & 0.70  & 0.74  & 2.96 \\
    \hline
    ENMAN303 & Course-specific & 0.4   & 1     & 23512 & 0.88  & \textbf{0.90} & 4.88 \\
    ENMAN303 & Course-specific & 0.4   & 2     & 18986 & \textbf{0.91} & 0.89  & 4.88 \\
    ENMAN303 & Course-specific & 0.4   & 4     & 17175 & 0.90  & 0.86  & \textbf{5.03} \\
    \hline
    ENMAN303 & Course-specific & 0.6   & 1     & 16750 & 0.87  & \textbf{0.90} & 4.88 \\
    ENMAN303 & Course-specific & 0.6   & 2     & 12001 & \textbf{0.90} & \textbf{0.90} & \textbf{5.25} \\
    ENMAN303 & Course-specific & 0.6   & 4     & 10501 & \textbf{0.90} & 0.86  & 5.03 \\
    \hline
    ENMAN303 & Course-specific & 0.8   & 1     & 10490 & 0.84  & \textbf{0.87} & 4.88 \\
    ENMAN303 & Course-specific & 0.8   & 2     & 6197  & \textbf{0.88} & 0.83  & 4.88 \\
    ENMAN303 & Course-specific & 0.8   & 4     & 5545  & 0.86  & 0.83  & \textbf{5.03} \\
    \hline
    
    \end{tabular}%

\end{table*}

\begin{table*}[!htp]
\caption{{Overview of the results for FCS-001 and all courses combined. All experiments are performed using SVM and random oversampling of the minority class. The metrics are assessed by using 10-fold cross validation. The best result per metric for a combination of course, pattern type and support is shown in bold.} }
\label{tab:all_courses}
\def\arraystretch{1}
\ignorespaces 
\centering 
\scriptsize
\begin{tabular}{llrrrrrr}
\hline
    Dataset & Patterns & \multicolumn{1}{l}{support} & \multicolumn{1}{l}{w} & \multicolumn{1}{l}{\#patterns} & \multicolumn{1}{l}{Acc} & \multicolumn{1}{l}{AUC} & \multicolumn{1}{l}{lift} \\
    \hline
    FCS-001 & Abstract & 0.4   & 1     & 70    & 0.59  & 0.52  & 1.29 \\
    FCS-001 & Abstract & 0.4   & 2     & 117   & \textbf{0.66} & \textbf{0.63} & \textbf{1.93} \\
    FCS-001 & Abstract & 0.4   & 4     & 115   & 0.65  & 0.58  & 1.72 \\
    \hline
    FCS-001 & Abstract & 0.6   & 1     & 46    & 0.59  & 0.51  & \textbf{1.72} \\
    FCS-001 & Abstract & 0.6   & 2     & 49    & \textbf{0.64} & 0.54  & 1.29 \\
    FCS-001 & Abstract & 0.6   & 4     & 65    & 0.61  & \textbf{0.57} & 1.50 \\
    \hline
    FCS-001 & Abstract & 0.8   & 1     & 20    & 0.50  & 0.38  & 0.21 \\
    FCS-001 & Abstract & 0.8   & 2     & 49    & 0.60  & 0.51  & 1.50 \\
    FCS-001 & Abstract & 0.8   & 4     & 29    & \textbf{0.67} & \textbf{0.56} & \textbf{1.72} \\
    \hline
    FCS-001 & Course-specific & 0.4   & 1     & 818   & 0.66  & \textbf{0.57} & \textbf{1.50} \\
    FCS-001 & Course-specific & 0.4   & 2     & 533   & 0.65  & 0.52  & 1.29 \\
    FCS-001 & Course-specific & 0.4   & 4     & 446   & \textbf{0.67} & 0.55  & 1.07 \\
    \hline
    FCS-001 & Course-specific & 0.6   & 1     & 617   & 0.66  & \textbf{0.57} & \textbf{2.15} \\
    FCS-001 & Course-specific & 0.6   & 2     & 284   & \textbf{0.61} & 0.54  & 1.29 \\
    FCS-001 & Course-specific & 0.6   & 4     & 232   & \textbf{0.61} & 0.54  & 0.86 \\
    \hline
    FCS-001 & Course-specific & 0.8   & 1     & 267   & \textbf{0.65} & \textbf{0.60} & \textbf{1.07} \\
    FCS-001 & Course-specific & 0.8   & 2     & 83    & \textbf{0.65} & 0.58  & \textbf{1.07} \\
    FCS-001 & Course-specific & 0.8   & 4     & 100   & 0.57  & 0.51  & 0.43 \\
    \hline
All courses & Abstract & 0.4   & 1     & 67    & \textbf{0.64} & \textbf{0.64} & \textbf{1.76} \\
    All courses & Abstract & 0.4   & 2     & 99    & 0.59  & 0.61  & 1.71 \\
    All courses & Abstract & 0.4   & 4     & 135   & 0.58  & 0.59  & 1.61 \\
    \hline
    All courses & Abstract & 0.6   & 1     & 84    & \textbf{0.62} & 0.60  & 1.70 \\
    All courses & Abstract & 0.6   & 2     & 76    & 0.57  & 0.60  & 1.54 \\
    All courses & Abstract & 0.6   & 4     & 43    & 0.61  & \textbf{0.61} & \textbf{1.83} \\
    \hline
    All courses & Abstract & 0.8   & 1     & 36    & 0.65  & 0.60  & 1.21 \\
    All courses & Abstract & 0.8   & 2     & 14    & \textbf{0.68} & \textbf{0.60} & \textbf{1.43} \\
    All courses & Abstract & 0.8   & 4     & 10    & 0.52  & 0.52  & 1.00 \\
    \hline
    
    \end{tabular}%

\end{table*}

A few additional trends can be observed. First, the models built based on course-specific patterns outperform general patterns with an average AUC score of 0.7 and a lift of 3.22 compared to an average AUC equal to 0.63 and lift of 2.43, respectively. This observation is not entirely unexpected; however, what is interesting is that the performance of the general patterns-based models is still relatively good, especially if the complexity of the blended learning setting is considered. For instance, the performance metrics for the models built for \textit{ENFGF243} reach an accuracy of 0.76, AUC of 0.71 and lift of 3.24, which means that using more coarse-granular patterns can sometimes be sufficient for predicting student success. It can be concluded that a larger alphabet size with more detailed information will most likely lead to better predictions; however, if generalisability of the results is important, the coarser level of granularity could be used as well, even at the levels when no information about the tackled object is known and only the action information is provided to the algorithm.

Similarly to the results based on general patterns for the other courses, the predictive models for all courses combined do not achieve very strong predictive performance, but still are able to outperform a random model with the highest AUC equal to 0.64. One of the reasons for the complexity of predicting students scoring in the bottom 20\% is the differences in scoring across different courses. 

Moreover, as can be seen in Table \ref{tab:eol_overview}, the score distributions for some courses are characterized by low variability, with e.g., the lowest scores of 5 and the highest score of 6. Given such low variability, the behavior of students whose score is in the bottom 20\%, might not differ sufficiently from those obtaining a higher score. Naturally, predicting groups with more distinct behavior is a less challenging task, and therefore, for courses with a higher number of low-scoring cases, the prediction performance of sequence classification models based on general patterns will likely be higher.

The models with support = 0.6 achieve the best results for all three metrics analyzed (with $accuracy = 0.72$, $AUC = 0.66$, and $lift=2.77$ on average). The lower the support, the more patterns are discovered by iBCM, which leads to better predictions when support is lowered from 80\% to 60\% as more patterns are correlated with better predictions. Conversely, when support is further lowered to 0.4, more of the discovered patterns become irrelevant, and the predictive performance decreases.

The influence of time windows on predictive performance can be observed in Tables \ref{tab:results} and \ref{tab:all_courses}. In two-thirds (14 out of 21) of the combinations of the course, pattern type and support, the performance metrics are higher for the cases of 2 and 4 windows when compared with 1 window. For example, for \textit{ENMAN303} with general patterns, the AUC score increases from roughly 0.6 to 0.8 when the number of windows is increased from 1 to 2. However, this trend is not present for all the cases. For example, for the \textit{all courses} dataset, increasing the number of windows influences the predictive performance negatively. This can be attributed to the fact that the timing information becomes less meaningful and less related to the course content when many courses with different timelines are combined.

Another result that can be derived from the experimental results is that, while it is typical for the two-windows setup to outperform the one-window setup, the models with four windows tend to perform worse than those with two windows. There are several factors that can play a role in such a finding. First, it is to be expected that temporal information allows to grasp more specific and informative patterns when compared to when no windows are used. However, the way the windows were defined in the current set of experiments - i.e., by dividing the course timeline in a set of equal time intervals - might not correspond to a meaningful time division within a course. Mining patterns based on when the course content was released might be more beneficial in this context. Second, it is important to note that discovering frequent patterns in the whole sequence is easier for iBCM when compared to mining patterns in a subsequence limited to a certain window. As such, lowering the support level might allow to ease the requirements for the pattern to be considered frequent, and could facilitate discovering more patterns in the time-based subsequences.

\subsection{Pattern interpretation}

\begin{figure}
	\centering
		\includegraphics[width=0.9\textwidth]{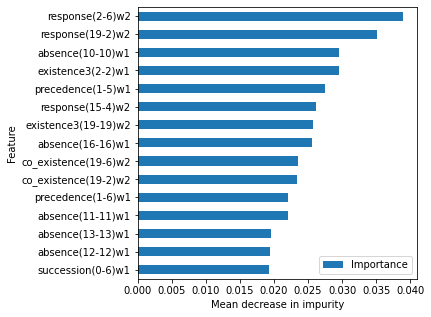}
		\caption{Top 15 most important sequential patterns found by iBCM based on mean decreased impurity of the best-performing Random Forest model built for the ENMAN303 dataset. The interpretation of the events is as follows:\\
		$0 - enrollment.activated$,\\
		$1 - course.upgrade.displayed$,\\
		$2 - ui.lms.link\_clicked$,\\
		$4 - edx.ui.lms.sequence.next_selected$,\\
		$5 - load\_video$,\\
		$6 - page\_close$,\\
		$10 - stop\_video$,\\
		$11 - problem\_check$,\\
		$12 - problem\_submitted$,\\
		$13 - problem\_graded$,\\
		$15 - ui.lms.sequence.previous\_selected$,\\
$16 - ui.lms.sequence.tab\_selected$,\\
$19 - openassessment.upload\_file$.\\
		}
		\label{fig:feature_importance}
\end{figure}

As shown above, sequence classification techniques are capable of discovering a large number - sometimes thousands - of meaningful patterns, which can be used to built accurate predictive models. However, given a large number of generated patterns, a possible usage of feature space transformation techniques and the fact that classification algorithms often serve as black boxes, it is not always easy to interpret the results of the models at the pattern level. Moreover, it is important to be cautious when interpreting causal relationship between patterns and student performance, since correlation between a desirable learning outcome and the pattern does not always imply that following this pattern is recommended. Nevertheless, relative importance of sequential patterns in a predictive model could provide the educators with an indication towards which student actions might have had the largest influence to the student performance, and a direction for further investigation of the possible reasons of this influence. Ideally, such an investigation could lead to either supporting the discovered patterns by existing educational theories, or to emerging of new educational theories.  

This section provides examples of important sequential patterns. Since it is even more conceptually interesting to investigate which abstract patterns are important, as it provides more opportunities to generalize, we illustrate the types of sequential patterns based on the results of the best-performing model for the abstract pattern types, which is the model built for ENMAN303, with support equal to 0.8 and with 2 time windows.

As discussed in Section \ref{sec:background}, iBCM is in essence a feature engineering technique, which extracts binary features from learners' activity sequences, and as such, it can be combined with any classification algorithm, such as SVM, random forest, logistic regression, etc. Although we use SVM as the models for calculating the results in the previous subsection due to its overall higher performance with sequential patterns, random forest can provide a better tool for visualizing feature importance in a machine learning model. Therefore, in this subsection, random forest is used for illustration of important patterns. The AUC of the models built by random forest and SVM has been equal to 0.8 for both algorithms for the selected case (ENMAN303, abstract patterns, support of 0.8, and w2).

The 15 most important sequential patterns discovered by iBCM are shown in Figure \ref{fig:feature_importance} (the interpretation of the numerical encoding of the events is provided in the caption of the figure). Given the nature of a random forest model, gaining a full understanding of the decision process based on this features is infeasible. Nevertheless, the potential meaning of why these behavioral patterns are important for student classification is quite intuitive. For instance, it is possible to draw a conclusion that participation in the intermediate quizzes is an important milestone and is predictive of the final outcome. It is also indicative not only whether the student participated in a quiz (here called problem or assessment), but also whether it was followed by a closed page, implying that the student exited the course (see e.g. pattern $co-existence(19-6)w2$), or by another quiz. For instance, the pattern $existence3(19-19)w2$ reveals whether the assessment has been uploaded over 6 times in the second half of the course, which is most likely negatively correlated with scoring in the bottom 20\% of the course. 

The occurrence of the patterns in the first half of the course (w1) and the second part of the course (w2) goes in line with the general intuition as well. Based on the patterns, it can be assumed that smaller quizzes (codes 11-13) played an important role in the first half of the course, whereas the participation in the assessment (19) has been important in the second part of the course. An ability to pinpoint student participation in the course activities to a specific time window of the course can provide course instructors with an opportunity to gain better understanding of how the course is followed and perceived. Conversely, additional information gained from the teachers, such as which patterns to search for in which window, could help to develop better and more aware predictive algorithms.

\section{Discussion}
\label{sec:discussion}

The method for capturing temporal aspects in behavioural sequences proposed in this paper has been shown to positively influence the accuracy of predicting student success. However, this influence is not present for every case and is not always very pronounced. This can be attributed to several reasons, such as too short sequences, too high support values, as well as the fact that the time windows have been identified by dividing the course timeline into equal time intervals rather than by considering the course information about when certain content items are released. More experiments in the future could be beneficial to investigate an optimal method for using time windows with iBCM, capable of elevating model performance even further.

Nonetheless, irrespectively of whether an increase in performance metrics is present for temporal windows, being able to pinpoint the exact timing of a pattern in a sequence is in itself an interesting feature. Such functionality can provide better interpretation of the results and facilitate the translation of the discovered patterns to the course context. Moreover, if the knowledge of the teachers about the desired pattern timing is considered, the time window functionality could be used to observe whether the students follow the learning path foreseen by the teacher.

When it comes to event granularity, a larger alphabet size containing more information about the student actions will most likely improve the predictive performance of the models. Notwithstanding, this not only comes at the price of the results’ generalisability, but also requires more computational time, as the running times of the experiments have been increasing proportionally to the number of sequential patterns discovered. Besides, when a 10-fold cross validation setup is utilised, the presence of such an increased number of patterns (e.g. over 10,000 patterns for the \textit{ENMAN303} models) need to be discovered for each fold separately, which is computationally even more expensive. 

\section{Conclusion and future work}
\label{sec:conclusion}

This paper has demonstrated the construction of predictive models for detecting underperforming students based on behavioural data. Hereto, sequence classification is used to derive behavioural patterns correlated with student performance, which can be transformed into recommendations for improving the quality of online education at UChile and beyond. The possibility to identify the students that will score in the bottom 20\% of the class will allow teachers to potentially target those students providing better support and offer them additional explanations that could improve their learning outcomes.

The results reveal that, although a larger alphabet size leads to discovering more informative pattern, more general non-course specific information could be used for constructing predictive models as well. In addition, the utility of the proposed time window functionality is discussed in terms of both predictive performance as well as explainability. Future research could focus on exploring this topic in more depth, by comparing different windowing setups in iBCM. For example, the performance of the equifrequent vs. equidistant approach can be compared. Moreover, different ways of dividing the course timeline could be investigated.

\section*{Acknowledgements}
RW gratefully acknowledges financial support from ANID PIA/BASAL\\ 
AFB180003, Fondecyt 1181036, Fondecyt 1221562, and FONDEF ID18I10216: "Desarrollo de tecnologías de big data para aumentar la retención y el éxito de estudiantes universitarios".

All authors acknowledge financial support from NeEDS Secondments “Research and Innovation Staff Exchange Network of European Data Scientists” (NeEDS); H2020-MSCA-RISE-2018, Project ID 822214.

\bibliography{main}

\end{document}